\newcommand{\Neu}[1]{#1}
\documentstyle[11pt,twoside,dm]{article}
\title{ 
Basic Formal Properties of A Relational Model of The          
Mathematical  Theory of Evidence
} 
\shorttitle{Properties of a MTE relational model}

\author{ Mieczys{\l}aw A. K{\l}opotek, S{\l}awomir T. Wierzcho\'{n}}
\address{
 Institute of Computer Science \\ Polish Academy of Sciences \\ Warszawa,  
Poland \\
e-mail: klopotek,stw@ipipan.waw.pl}
\newcommand{\rDef}[1]{ Def.\ref{#1}}
                       \newcommand{\SQL}[1]{ 
\noindent {
\it\begin{center}
\begin{tabular}{p{12cm}}
 #1
\end{tabular}
\end{center}
 }
\noindent
}

                \newcommand{\SQLinline}[1]{ {\it #1 } }

\newcommand{\izero}{\mbox{\bf I}} 
\newcommand{\ione}{\mbox{\bf I1}} 
\newcommand{\itwo}{\mbox{\bf I2}} 
\newcommand{\iione}{\mbox{\tiny\bf I1}} 
\newcommand{\iitwo}{\mbox{\tiny\bf I2}} 

\begin{document}

\machetitel

\begin{abstract} 
The paper presents a novel view of the Dempster-Shafer belief function as a 
measure of diversity in relational data bases.
The Dempster rule of evidence combination corresponds to 
the join operator of the relational database theory.
This rough-set based interpretation is qualitative in nature 
 and can represent a number of  belief function operators. 

\footnote{
I.2.3, I.2.4. 
}
\end{abstract}


\section{Introduction}
\newcommand{\DST}{MTE }
\newcommand{\DSTNB}{MTE}

Belief functions and their mathematical properties were investigated by A.P. 
Dempster in a series of papers in the late sixties \cite{Dempster:67}. They 
were intended as a generalisation of Bayesian inference in the sampling 
 situation. 
In the mid-seventies G. Shafer \cite{Shafer:76} (who coined the term belief 
 functions) 
developed a set of formal tools for the representation and combination of 
 evidence.
His aim was to construct a rather general theory, termed Mathematical Theory 
 of Evidence,
or MTE for short, for coping with uncertain but non-probabilistic 
 information. 
In spite of its numerous interesting formal properties, the theory caused 
great discussion centred around the validity of the axioms and manageable
interpretation of a belief function - see e.g. \cite{%
Dempster:67,%
Fagin:91,%
Halpern:92,%
Hummel:88,%
Kyburg:87,%
Ma:91,%
Pearl:90,%
Provan:90,%
Shafer:90,%
Skowron:94%
}.
 
Dempster \cite{Dempster:67} initiated interval interpretation of MTE, which 
 in 
fact is compatible with the random set theory \cite{Wierzchon:88}.  
H.~Kyburg \cite{Kyburg:87} showed  that the belief function may be 
 represented
by an envelope of a family of probability functions and claimed that the 
behaviour of combining evidence via belief functions may be properly
explained in statistics under proper independence assumptions.
Hummel and  Landy \cite{Hummel:88} considered MTE as a "statistics of expert
opinions" so that it "contains nothing more than Bayes' formula applied
to Boolean assertions, \dots (and) tracks multiple opinion as opposed to a
single probabilistic assessment". Pearl \cite{Pearl:90} and Provan 
\cite{Provan:90} considered belief functions as "probabilities of 
 provability". 
Still another view has been developed in connection with rough set theory
\cite{Grzymala:91,%
Skowron:93,%
Skowron:93b,%
Skowron:94}. 
Belief function is considered there as the
lower approximation of the set of possible decisions in a (partial) decision
table.
Fagin and  Halpern \cite{Fagin:91} postulated probabilistic interpretation of
MTE around lower and upper probability measures defined over a probability
structure (rather than space).
The list of other attempts is quite long. 

Though a tendency to consider belief functions as subjective uncertainty 
measures is visible \cite{Shafer:90ijar}, the need  for case-based 
interpretation as a pre-condition for practical applicability has been 
explicitly stressed  \cite{Wasserman:92ijar}. Still this interpretation 
should be qualitative rather than quantitative in nature \cite{Smets:92}. 
However, these requirements seem not to be met so far. 

Apart from manageable interpretation, the MTE caused troubles of purely 
numerical nature. Unlike ordinary probabilities, which assign mass to each 
possible outcome, belief functions assign mass to each subset of the outcome 
 space. 
As a consequence the amount of memory space required to store a belief 
 function
in a computer will grow exponentially with the size of the outcome space we
consider. Searching for economical representation of a belief function, the
researches made use of the already known concepts of sets factorisation 
 analysed 
in the theory of relational databases. 
In our search for non-quantitative interpretation of MTE, our attention was 
attracted by the nature of the join operator of relational databases
\cite{Beeri:83}  or in general the multivalued dependency \cite{Delobel:78}
the study of which led to invention of local computation method for 
 uncertainty 
propagation of Shenoy and Shafer \cite{Shenoy:94} for MTE.  This in turn 
 made 
the rough-set theoretic interpretation of MTE belief functions of Skowron 
 and Busse \cite{Skowron:94} best choice for 
further investigation, as it was purely-case based and relational, though it 
is frequency based. The rough-set interpretation sheds some light onto 
what the  concept of "evidence" may mean in experimental terms.  The 
"evidence" there is the information part of a database record and it 
 "supports" 
the decision part of a record. The complexity of combination of evidence 
according to the Dempster rule in \cite{Skowron:94} gave an impulse for 
search of a simpler way to accomplish it. It turns out that with frequency  
approach updating of decision parts of cases is needed 
(Dempster's combination is destructive). Out of this experience we decided 
to abandon the frequencies and concentrated on purely relational operations. %


The paper is organised as follows: Section 2 briefly introduces basic \DST 
concepts. Section 3 recalls traditional rough set theoretic frequency 
interpretation of \DST from \cite{Skowron:94} and explains our insights of 
destructive nature of Dempster's combination with respect to frequencies. 
Section 4 presents our new qualitative interpretation.  
The paper ends with some concluding remarks. 

Throughout the paper, as relational data tables are subject of rough set 
theory, SQL \cite{SQL} query language is used to express semantics of \DST 
measures and operators in terms of decision tables, both with respect to 
traditional and our rough sets based interpretation  of \DSTNB, as SQL has 
 the 
capability of expressing purely relational and frequentistic data processing.

\section{Basics of the Dempster-Shafer Theory}

We understand \DST measures in a very traditional way
(see \cite{Provan:90}). 
Let $\Xi$ be a finite  set of elements called elementary events. 
Any subset of $\Xi$ is a composite event, or hypothesis. $\Xi$ be called also the 
frame of discernment. 
A basic probability (or belief) assignment (bpa) function is any set function 
m:$2^\Xi
\rightarrow [0, 1]$ such that  $$  \sum_{A \in 2^\Xi } m(A)=1   
\qquad
  m(\emptyset)=0, \qquad 
\forall_{A \in 2^\Xi} \quad  0 \leq   m(A)$$
We say that a bpa             is vacuous iff $m(\Xi)=1$ and $m(A)=0$ for every 
$A\ne\Xi$. 
A belief function is defined as Bel:$2^\Xi \rightarrow [0,1]$ so that 
 $Bel(A) = \sum_{B \subseteq A} m(B)$.
A plausibility function is Pl:$2^\Xi \rightarrow [ 0,1]$  with 
$\forall_{A \in 2^\Xi}   Pl(A) = 1-Bel(\Xi-A )$.
A commonality function is Q:$2^\Xi-\{\emptyset\} \rightarrow [0,1]$ with 
 $\forall_{A \in 2^\Xi-\{\emptyset\}} \quad Q(A) = \sum_{A \subseteq B}
m(B)$. 

The     Rule-of-Combination of two Independent Belief Functions 
$Bel_{E_1}$, $Bel_{E_2}$ 
Over-the-Same-Frame-of-Discernment (the so-called Dempster-Rule), 
 denoted 
    $Bel_{E_1,E_2}=Bel_{E_1} \oplus Bel_{E_2}$ 
 is defined interms of bpa's as follows: 
$m_{E_1,E_2}(A)=c \cdot  \sum_{B,C; A= B \cap C} m_{E_1}(B) \cdot 
m_{E_2}(C)$ (c - constant normalizing the sum of $m$ to 1).

Under multivariate settings $\Xi$ is a set of vectors in n-dimensional space 
spanned by the set of variables {\bf X}=\{ $X_1, X_2, \dots X_n$\}. 
If $A\subseteq\Xi$, then by projection  $A^{\downarrow {\bf Y}}$ 
of the set  $A$ onto a subspace spanned 
by the set of variables ${\bf Y}\subseteq {\bf X}$ we 
understand the set $B$ of vectors from $A$ projected onto {\bf Y}.   
Then marginalization operator of \DST is defined as follows: \linebreak
$m ^{\downarrow {\bf Y} }(B)= 
\sum_{A; B=A  ^{\downarrow {\tiny
X} }} m(A)$.

\begin{definition} \label{ShaferCond}
 (See \cite{Shafer:90b}) Let B be a subset of $\Xi$, called 
evidence,
 $m_B$ be a basic probability assignment such that $m_B(B)=1$ and $m_B(A)=0$
for any A different from B. Then the conditional belief function $Bel(.||B)$
representing the belief function $Bel$ conditioned on evidence  B 
is defined
as: $Bel(.||B)=Bel \oplus Bel_B$. 
\end{definition}

\section{Rough Set Theory. The Traditional Interpretation of Belief 
 Functions}

Skowron and Grzymala-Busse \cite{Skowron:94} and others studying rough
sets developed more
specifically the proposal of Shafer with respect to frequency interpretation
of \DSTNB.

\Neu{%
Let us introduce the following denotation concerning decision tables.
 Let a tuple $\mu$ mean a function $\mu:A\rightarrow 
DOM(A)$, with $A$ being a set of attributes $A_j$, $DOM(A_j)$ being the 
domain of the attribute $A_j$, $DOM(A)=\bigcup_{A_j\in A} DOM(A_j)$.
$A$ be called the scheme of $\mu$, $A=S(\mu)$. A relational table $TAB$ be 
any set of tuples with identical scheme. This common scheme be denoted by 
$S(TAB)$. Let $\mu[R]$ with $R\subseteq S(\mu)$ denote the restriction of 
the tuple $\mu$ to the scheme $R$: 
$\mu[R]=\{(A_j,a_{jk}) | A_j\in R \land a_{jk}=\mu(A_j)\}$. 
The restriction of a relational table $TAB$ to $R$, denoted $TAB[R]$, 
be defined $TAB[R]=\{\mu[R] | \mu \in TAB\}$.
A relational join of two relational tables $TAB_1,TAB_2$ be defined as:
$TAB_1 \otimes TAB_2=\{\mu_1\cup\mu_2 | \mu_1\in TAB_1,
\mu_2\in TAB_2 \land
\forall_{A_j\in S(\mu_1)\cap S(\mu_2)} \  \mu_1(A_j)=\mu_2(A_j)\}$. 

A decision table is a relational table in which we 
split the scheme into  two distinct 
parts: the information part {\bf I } and the decision part  {\bf D}.

Let $card(SET)$ denote the cardinality of the set SET. 

}

  Let us assume that a decision table $TAB$ of decisions $D$ (atomic 
values)  under 
conditions (information) {\bf I } 
(atomic value vectors)
is available. 
{However, {\bf I } may not contain the complete information to make 
 decision $D$. 
} %
This gives rise in a natural way to a   mapping $\Gamma$
 assigning different
values of $D$ to the same value of {\bf I}. Under these circumstances the
belief  
in a set $A$ (subset of the domain of $D$) $Bel(A)$ may be derived from a case 
database as follows:
 $Bel(A)=1- card(\{\mu | \mu \in TAB \land \mu(D)\not\in 
A  \}) /card(TAB)$, which may be implemented  
 the SQL query language \cite{SQL} as:
\SQL{create view Total (Counted) as select count(*) from TAB;\\
select count(*)/Counted from TAB,Total
where not ({\bf I } in 
(select {\bf I } from TAB where not (D in A)));
}
Skowron and Grzyma{\l}a-Busse\cite{Skowron:94}
elaborated also a notion of 
conditioning under rough set interpretation 
\cite[p.219 ff.]{Skowron:94} as
respective measures for 
subtables (that is tables 
consisting of cases selected by a criterion).  Let us condition on $D$
belonging to the set $B$ selecting tuples fitting the condition $B$ 
$Subtable=\{\mu | \mu \in TAB \land \mu(D)\in B$
which may be implemented  as 
\SQL{create view SubtableTAB(D,{\bf I }) as
select X,{\bf I }              from TAB
where not ({\bf I } in
(select {\bf I } from TAB   where not (D in B)));
}
The belief distribution for the subtable can be calculated
from the view SubtableTAB in the same way as for the TAB database.
However, it
is a matter of a calculation exercise to show that their notion of
conditionality does not agree with that of Shafer from \rDef{ShaferCond}. 

Therefore, to achieve
consistency with Shafer's conditioning from \rDef{ShaferCond} we propose    
the following interpretation, 
derivable from our approach described in the paper \cite{Klopotek:95i}:
Let $v \in B$. Then in SQL 
\SQL{update TAB set D=v where not (D in B)
and {\bf I } in (select {\bf I } from TAB where D in B);\\
delete TAB where not (D in B);
}
Let $c_1$,$c_2$ be two cases from the database TAB such that ${\bf I 
}(c_1)={\bf I }(c_2)$
but  $D(c_1)\ne D(c_2)$ Let be $D(c_1)\in B$ and  $D(c_2)\not\in B$ 
Then after the above update and delete operations both cases $c_1,c_2$ are
retained in the database TAB. $c_1$ has retained its value $D(c_1)$. But the
case $c_2$ was subject to a metamorphose: its $D(c_2)$ has been changed (to
v). This means that the Dempster rule
of combination is "destructive". Preservation of frequency interpretation
under conditioning enforces ignoring the intrinsic (observed) value of an
attribute and replacement of it with some other value. 
It should be stressed at this point that the above SQL operation cannot be easily expressed in terms of sets and relations, because an update operation is engaged which may make distinct tuples identical.  

A still more complex task is the interpretation of combination of two 
independent pieces of evidence. Skowron and Grzyma{\l}a-Busse\cite{Skowron:94} 
 elaborated a procedure consisting in transforming the combined decision 
tables 
into a kind of summary with multivalued decision columns (since 
non-relational) and  then applying complex rational arithmetic to get 
finally 
a decision table 
(derived by so-called $\Psi$-independent combination) implementing 
   Dempster 
combination of independent belief functions
(consult  \cite{Skowron:94} for 
details). 

\section{New Interpretation}

Below we present a slight modification of the rough set interpretation of 
belief functions that surprisingly turns out to be both simple,    elegant 
and 
straight forward and at the same time fulfils the requirement that was not 
matched by any known interpretations: it is qualitative and not quantitative 
in nature and still case-based. 
Furthermore we demonstrate that our new interpretation corresponds strictly 
to the notion of multivalued dependency, that is combination of belief 
functions parallels the relational join operator from database theory. 

The results of any  experiment with multiple outcomes
 can evaluated along two dimensions: 
the quantitative and the qualitative one. If we say e.g.
that in a series of coin tossing experiments we got 57 heads and 43 tails, then this is a quantitative evaluation. But if we say that there were heads and tails (and not e.g. edges)  then we say something about the qualitative aspects of the experiment. In the quantitative evaluation we say 
that 57 \% of all the cases we got heads, 
in  the qualitative evaluation we say 
that 50 \% of all possibilities of diverse outcomes are heads. 
In most real life cases we are more interested in the quantitative aspects.
Sometimes, however, the qualitative side may be of more interest.
E.g. if 40 witnesses say that they saw the suspect had killed the victim,
but their testimonies are suspiciously similar,  
and 20 say they saw the contrary, and their testimonies 
made impressions of individuality, then we would say that there are 1:20 chances of the guilt of the suspect rather than 40:20, because the qualitative aspects (diversity) 
 are more important than quantitative ones (frequency). 

Behaviour of frequencies under reasoning was historically the foundation of probability theory. As probabilistic (frequency based) models          of 
MTE fail in general, we considered just the diversity  as a possible alternative for a model of MTE. The diversity is well handled by relational model of databases. Though at a first glance the count of cases
in a relational database may appear identical with counting frequencies of objects / events, the difference starts as soon as we make a projection on the subset of attributes. It turns out that projected frequencies 
differ significantly from the counts of cases in a projected relational database.  Therefore we say that our approach is non-frequency, non-quantitative, that is qualitative one.

Let us define the plausibility $Pl_{TAB}(SET)$ derived from a decision table 
TAB with decision variable D and the set {\bf I} of information variables as: 
$Pl_{TAB}(SET)=  card(\{\mu[\izero] | \mu \in TAB \land \mu(D)\in SET\})
 /card(TAB[\izero])$, 
implemented as
%
\SQL{create view tmpTAB(No) as select count(distinct {\bf I}) from TAB;\\
     create view plTAB as select count(distinct {\bf I})/No from TAB,tmpTAB 
where TAB.D in SET;} 

\Neu{%
Example 1 explains the detailed numerical procedure for calculation of $Pl$ 
from the above SQL expression.
}

%
\begin{theorem} \label{th0}
The function  $Pl_{TAB}(SET)$ derived from a decision table 
TAB with decision variable D and the set {\bf I} of information variables
is plausibility function $Pl(SET)$ in the sense of Dempster-Shafer theory. 
\end{theorem}
\AnfBeweis
In \DSTNB, the plausibility of a set SET is just the sum of basic probability 
assignments $m(A)$  such that $SET \cap A \ne \emptyset$. 
Let $r$ be a record, $I(r)$ its information part and $D(r)$ its decision 
 part. 
For the set SET, 
let us consider a subset $R$  of all records from the decision table TAB such 
that: if $r\in R$ then $D(r)\in SET$ 
and for every $d \in SET$ there exists $r'\in R$ such that $D(r)=D(r')$ and 
there exists no record $r"\not\in R$  such  that $I(r")=I(r)$.
Obviously, for two distinct sets $SET'$ and $SET"$ their respective sets $R'$ 
and $R"$  
will share no records. Furthermore, $\bigcup_{SET\subseteq domain(D)} R(SET)$ 
will be the (relationally) identical with TAB. Hence we can consider 
the ratio number of records with distinct information part in $R(SET)$ 
 divided 
by the number of records with distinct information parts in DT as the bpa 
function $m(SET)$ in the sense of \DSTNB.  
But the function 
$Pl_{TAB}(SET)$ counts (the relative share of) the records with distinct 
information part such that the decision part belongs to SET. Hence 
in practice 
it is just the sum of basic probability assignments $m(A)$  such that $SET 
\cap A \ne \emptyset$. Therefore it is a plausibility function. 
\EndBeweis
\begin{example}
Let us first look at the  relational table 
BUILD in tab.  \ref{t1}). The 
column D is the 
decision column, I                 is  information part of the table. 
The domain of the decision variable D is \{center, restaurant, school\}.

\begin{table}
\caption{
Example: Public offering: erection of buildings of a school, a restaurant and 
a shopping center. 
Decision table: BUILDing. 
"Information part" - the firm, "Decision part"  - the object to be erected
}
\label{t1}
\begin{center}
\begin{tabular}{| r | l || l |}
\hline
 & \multicolumn{1}{c |}{I} &  \multicolumn{1}{c |}{D}   \\
\hline
1.&ABD A.G.&center\\
2.&LQR Inc. & school\\
3.&PTS Ltd.&center\\
  &PTS Ltd.&restaurant\\
4.&XYZ Inc.&center\\
5.&ZZZ Ltd. & restaurant\\
  &ZZZ Ltd. & school\\
\hline
\end{tabular}
\end{center} 

\end{table}
%

Let us calculate now the plausibility 
  Pl(\{school, restaurant\})
from this table.
There are 7  cases (rows) in the dataset.  But there are only 5   cases with 
distinct 
information part (firms) {\bf I}.  
 And there are only 3   cases with decision either
 $school$ or  $restaurant$
with distinct 
information part {\bf I} (LQR Inc., PTS Ltd., ZZZ Ltd).
So the plausibility\footnote{%
Notice that under Skowron/Busse interpretation \cite{Skowron:94} 
we get $Pl= 4/ 7$ which is 
obviously a different value. The difference stems from the fundamental 
difference between frequency (Skowron/Busse) and relational(ours) view of the 
world.}
 is equal to Pl(\{school, restaurant\})=$3/5$.
One can check that 
Pl(\{school\})=2/5 (LQR Inc.,  ZZZ Ltd) and 
Pl(\{restaurant\})=2/5 (PTS Ltd.,  ZZZ Ltd). 
\end{example}

Notice that 
from the calculational rules for Dempster-Shafer theory we can derive also 
relational views calculating other measures:\\
Belief  
 $Bel_{TAB}(SET)=1- card(\{\mu[\izero] | \mu \in TAB \land \mu(D)\not\in 
SET\}) /card(TAB[\izero])$, implemented as 
%
%
\SQL{create view belTAB as select 1-count(distinct {\bf I})/No from 
TAB,tmpTAB 
where not (TAB.D  in SET);} 
Commonality 
 $Q_{TAB}(SET)=
 card(\{\mu[\izero] | 
\forall_{d\in SET}\  \mu[\izero]\cup\{(D,d)\} \in TAB\})$ %
$/card(TAB[\izero])$,   implemented as 
%
%
%
\SQL{create view tmp1TAB(CN) as select count(distinct D) from TAB \\
      where TAB.D in SET group by {\bf I} ;\\
     create view qTAB as select count(*)/No from tmpTAB,tmp1TAB 
      where CN=card(set);
{\rm (card() is a function counting the elements of the set passed as its 
argument) } 
}
Basic belief assignment  $m_{TAB}(SET)= 
 card(\{\mu[\izero] | 
\forall_{d\in SET} \  \mu[\izero]\cup\{(D,d)\} \in TAB\}
\land
\forall_{d\not\in SET}\  \mu[\izero]\cup\{(D,d)\} \not\in TAB\}
) 
/card(TAB[\izero])$,   implemented as 
%
%

\SQL{create view tmp11TAB({\bf I},D) as
  select TAB.{\bf I}, TAB.D+XX.D from TAB, TAB XX 
  where TAB.D in SET and XX.{\bf I}=TAB.{\bf I};\\
     create view tmp12TAB({\bf I},CN) as
select {\bf I},count(distinct D) from tmp11TAB group  by {\bf I}; \\
     create view m as count(*)/No from tmpTAB, tmp12TAB\\
 where CN=card(SET)*card(SET);}


\begin{theorem} \label{th01}
The functions  $Bel_{TAB}(SET)$, $Q_{TAB}(SET)$, $m_{TAB}(SET)$   
 derived from a decision table 
TAB with decision variable D and the set {\bf I} of information variables
are belief, commonality, basic probability/belief assignment 
 functions resp. $Bel(SET), Q(SET), m(SET)$ in the sense of
Dempster-Shafer theory. 
\end{theorem}

\begin{example} From table \ref{t1} we easily calculate that:\\
\underline{Commonality} Q(\{school, restaurant\}=1/5 (%
Number of firms ready to build either the school and the restaurant: 
 ZZZ Ltd). \\
\underline{Belief} Q(\{school, restaurant\}=2/5 
Number of firms ready to build nothing but  the school or the restaurant (LQR 
Inc.,  ZZZ Ltd)\\
\underline{bpa} - No of firms exactly offering erecting of: 
m(\{school, restaurant\}=1/5 (ZZZ Ltd),
m(\{restaurant\}=0   (none), 
m(\{school\}=1/5 (LQR Inc.)\\
\end{example}

\begin{table}
\caption{Shafer's Conditioning. Relational Interpretation.
Calculating $Bel(.|| \{school, restaurant\})$ as a $Bel$ from the table 
obtained by 
 \SQLinline{select I,D from BUILD where D=school or D= 
 restaurant} 
}
\label{tsc} 
\begin{center}
\begin{tabular}{| r | l || l |}
\hline
 & \multicolumn{1}{c |}{I} &  \multicolumn{1}{c |}{D}   \\
\hline
1.&LQR Inc. & school\\
2.&PTS Ltd.&restaurant\\
5.&ZZZ Ltd. & restaurant\\
  &ZZZ Ltd. & school\\
\hline
\end{tabular}
\end{center} 
\end{table}

%

\subsection{Conditioning as Selection of a Subtable}

Let us define now  the conditional belief function $Bel_{TAB}(.||B)$
representing the belief function $Bel$ conditioned on evidence  B 
as the belief function $Bel_{TAB\_B}(.)$ define over the view table 
\SQL{create view TAB\_B({\bf I},D) as 
select {\bf I},D from TAB where TAB.D in B;}
\begin{example}
In our example, for the relational table BUILD from tab. \ref{t1}, 
$Bel_{BUILD}(.||\{school, restaurant\})$ is      just
 $Bel$ calculated from 
the respective projection 
 \SQLinline{select I,D from BUILD where D=school or D= 
 restaurant} 
visible in  tab. \ref{tsc}.  
It is easily seen that 
$Pl_{BUILD}(\{restaurant\}|| \{school, restaurant\})=2/3$ (PTS\ Ltd.,ZZZ\
Ltd.) and 
$m_{BUILD}(\{restaurant\}|| \{school, restaurant\})=1/3$ (PTS\ Ltd.). 
\end{example}%
Our conditional belief function matches perfectly the Shafer's definition of 
 $Bel(.||B)$ cited above (\rDef{ShaferCond}).
Notice that under Skowron/Busse interpretation
(section 3), the matching of Shafer's 
conditionality definition had to be paid for with creating a physical copy of 
a relational table and 
updating it, whereas our notion works perfectly without any updates - only 
selection is used just as in probabilistic conditioning. 

\subsection{Combination as Relational Join}

\begin{table}
\caption{Public offering: equipment for buildings of a school, a restaurant 
and a shopping center. 
Decision table: EQUIPment 
"Information part" - the firm, "Decision part"  - the object to be equipped}
\label{tequip}
\begin{center}
\begin{tabular}{| r | l || l |}
\hline
 & \multicolumn{1}{c |}{I2} & \multicolumn{1}{c |}{D}  \\
\hline
1.&AAA GmbH&school\\
2.&BBB Ltd. &center\\
  &BBB Ltd.&restaurant\\
3.&CCC Inc.&center \\
  &CCC Inc.&restaurant\\
\hline
\end{tabular}
\end{center} 
\end{table}

%

Now let us discuss the most impressive property of the new interpretation:
the Dempster's rule of combination interpreted as relational join.

\begin{example} 
Let us consider the decision table  EQUIP (tab. \ref{tequip})
and BUILD (tab. \ref{t1}).
We want to combine independent evidence from both tables 
to support a decision. Let us assume that 
independence of evidence means that 
 no pair of firms (one from BUILD, one from EQUIP) 
refuse to cooperate on erecting  and equipping an object. 
How many pairs of firms do we have to  finish  a  set  of  objects 
mentioned in the offerings ? 
The answer lies in the relational table FINISH
 (tab. \ref{tcomb})
 obtained
as a relational join of BUILD and EQUIP (over the common 
column D) so that the new decision table has as its decision column D and as 
its information part I,I2: 
\SQL{ create table FINISH (I,I2,D);\\
     insert into table FINISH from\\
     select I,I2,D from BUILD, EQUIP where BUILD.D=EQUIP.D;
} 
Notice that in BUILD, there were 5 cases with distinct information part,
in EQUIP - 3, and in  BUEQ  there are only 10.
We have here  
$Pl_{FINISH}(\{school, restaurant\})=8/10$ and 
$Bel_{FINISH}(\{school, restaurant\})=3/10$

You can easily check that $Bel_{FINISH}=Bel_{BUILD}\oplus Bel_{EQUIP}$. 
\end{example} 
Generally, we can formulate the theorem:

\begin{theorem} \label{th1}
If the decision tables DT1(\ione,D) and DT2(\itwo,D) 
with non-over\-lap\-ping information parts \ione,\itwo\ 
are combined by relational join operation 
$DT1\otimes DT2$, implemented as 
\SQL{select \ione,\itwo,DT1.D from DT1,DT2 where DT1.D=DT2.D;}
yielding  table DT12(\izero,D) with  \izero=\ione$\cup$\itwo, then 
 $Bel_{DT12}=Bel_{DT1}\oplus Bel_{DT2}$. 
\end{theorem}

\AnfBeweis
This can be demonstrated by considering the "fate" of records counted   on 
calculation of $m_i$. If $R_1$ is the set of records counted when calculating 
$m_1(A)$ in DT1, and  if $R_2$ is the set of records counted when 
calculating $m_2(B)$ in DT2, then upon join only records 
$\mu=(\mu_1[\ione],\mu_2[\itwo],\mu_1[D])$ 
with $\mu_1 \in R_1$, $\mu_2 \in R_2$, $\mu_1[D]=\mu_2[D]\in A \cap B$ 
will be created, hence they will be counted in support of $m(A\cap B)$. 
Furthermore, their number will be exactly equal to the product 
of the number of distinct records in $R_1$ times the number of distinct
records in $R_2$, so that the Dempster formula will be matched perfectly upon 
normalization. 
\EndBeweis
%
%
%

\begin{table}
\caption{Combination of Independent Evidence. Decision table FINISH
obtained as 
\SQLinline{select I,I2,D from BUILD, EQUIP where BUILD.D=EQUIP.D}
} 
\label{tcomb}
\begin{center}
\begin{tabular}{| r | l | l || l |}
\hline
 & \multicolumn{1}{c |}{I}  & \multicolumn{1}{c |}{I2} &  \multicolumn{1}{c
|}{D}   \\
\hline
1.&ABD A.G.&BBB Ltd. &center\\
2.&ABD A.G.&CCC Inc. &center\\
3.&LQR Inc. &AAA GmbH& school\\
4.&PTS Ltd.&BBB Ltd. &center\\
  &PTS Ltd.&BBB Ltd. &restaurant\\
5.&PTS Ltd.&CCC Inc. &center\\
  &PTS Ltd.&CCC Inc. &restaurant\\
6.&XYZ Inc.&BBB Ltd. &center\\
7.&XYZ Inc.&CCC Inc. &center\\
8.&ZZZ Ltd. &BBB Ltd. & restaurant\\
9.&ZZZ Ltd. &CCC Inc. & restaurant\\
10.&ZZZ Ltd. &AAA GmbH& school\\
\hline
\end{tabular}
\end{center} 
\end{table}


\subsection{Relational Marginalization and Decombination}

A further intriguing property, not present in any interpretation of MTE known so far, is the relationship between relational marginalization
and MTE factorization ("decombination") of belief functions. 

\begin{example}
Notice that BUILD and EQUIP in our example are both in first normal form and the 
 domain of the attribute D is identical in both tables. 
Therefore we know from elementary properties of relational data tables that 
 marginalization of FINISH over I,D            
\SQL{select distinct   I,D from FINISH;}
is exactly identical with BUILD. 
 and  marginalization of FINISH over I2,D
\SQL{select distinct   I2,D from FINISH;}
is exactly identical with EQUIP. 
\end{example}

In general:
%
%
\begin{theorem} \label{th2}
If 
the information part {\bf I } of 
the decision table DT(\izero,D) 
can be split into two such parts \ione,\itwo \  that 
$\ione\cup \itwo=\izero$ and $\ione \cap \itwo=\emptyset$ and 
the relation DT is identical with DT1$\otimes$DT2, implemented
\SQL{select \ione,\itwo,DT1.D from DT1,DT2 where DT1.D=DT2.D;}
where DT1 and DT2 are DT1=DT[\ione,D], DT2=DT[\itwo,D], implemented
\SQL{create view DT1 as select distinct   \ione,D from DT; \\
     create view DT2 as select distinct   \itwo,D from DT;} 
that is there is a multivariate dependency between \ione \ and \itwo \ given 
 D, 
then 
 $Bel_{DT}=Bel_{DT^{\downarrow \iione,D}}\oplus Bel_{DT^{\downarrow \iitwo,
D}}$. 
\end{theorem}
\AnfBeweis
Follows directly from theorem \ref{th1}.
\EndBeweis
Let consider the unnormalized \DST measures of decision tables 
$m'_{TAB}$, $Bel'_{TAB}$, $Pl'_{TAB}$, $Q'_{TAB}$, 
such that $f'_{TAB}=f_{TAB}\cdot card(TAB^{\downarrow I})$ (card - number of 
distinct rows, $f$ - $m$ or $Bel$ or $Pl$ or $Q$) and the unnormalized 
combination operator $\oplus'$ such that 
    $Bel'_{E_1,E_2}=Bel'_{E_1} \oplus' Bel'_{E_2}$ 
 is defined as follows: 
$m'_{E_1,E_2}(A)= \sum_{B,C; A= B \cap C} m'_{E_1}(B) \cdot 
m'_{E_2}(C)$.

What may be more surprising, a kind of a reverse theorem holds: 
\begin{theorem} \label{th3}
The information part I of 
the decision table DT(I,D) 
can be split into two such parts \ione,\itwo \ that 
$\ione\cup \itwo=I$ and $\ione \cap \itwo=\emptyset$ 
and 
 $Bel'_{DT}=Bel'_{DT^{\downarrow \iione,D}}\oplus' Bel'_{DT^{\downarrow 
\iitwo, D}}$ 
if and only if
the relation DT is identical with DT1$\otimes$DT2, implemented
\SQL{select \ione,\itwo,DT1.D from DT1,DT2 where DT1.D=DT2.D;}
where DT1 and DT2 are  DT1=DT[\ione,D], DT2=DT[\itwo,D], implemented
\SQL{create view DT1 as select distinct   \ione,D from DT; \\
     create view DT2 as select distinct   \itwo,D from DT;} 
that is there is a multivariate dependency between \ione \ and \itwo \ given 
 D, 
\end{theorem}
\AnfBeweis
(An outline.)
The if-part parallels exactly theorem \ref{th2}. We need only pay attention 
 to 
the fact that we never normalize.\\
The only-if-part follows from numerical calculations for Q-values of all the 
tables considered. 
If a record is counted in DT when calculating $Q'_{DT}(SET)$, then 
it is also counted when calculating both $Q'_{DT1}(SET)$ and $Q'_{DT2}(SET)$.
If we form a join $DT1\cdot DT2$ then  $Q'_{DT1\cdot DT2}(SET)=     
Q'_{DT1}(SET) \cdot Q'_{DT2}(SET)$. And this is the maximum value Q' can take 
in DT12. If there is ANY deviation from multivariate dependency concerning 
records with decision part in SET, then  value 
of $Q'_{DT12}(SET)$
is smaller than  $Q'_{DT1}(SET) \cdot Q'_{DT2}(SET)$. This proves our claim.
\EndBeweis
%
%
%
{\it Remak:}
We can conclude that Dempster's rule of combination is equivalent with 
relational join and the Dempster-Shafer independence of evidence means 
multivalued dependence of evidence.  
We can also simulate other rules of combination of evidence. 
In the above, we assumed that given the decision, we cannot conclude from the 
information part 
\ione\  the value of the information part 
\itwo\  in DT. This meant 
qualitative independence. Now let us assume the contrary in another decision 
table DT': given the 
decision d, we can totally predict 
\itwo\  from \ione\  for all records r with D(r)=d in 
DT' or  we can totally  predict 
\ione\  from \itwo\  for all records r with D(r)=d in 
DT'. It is immediately clear that in this case for any set of decisions
the unnormalized plausibility is calculated as 
$Pl'_{DT'}(A)=max(Pl'_{{DT'}^{\downarrow \iione,D}}(A),
Pl'_{{DT'}^{\downarrow \iitwo,D}}(A))$. 
We can conclude for normalized plausibility that we deal here with the know 
rule of combination of dependent evidence:  
$Pl'_{DT'}(A)=max(\alpha \cdot Pl'_{{DT'}^{\downarrow \iione,D}}(A),
(1-\alpha)\cdot Pl'_{{DT'}^{\downarrow \iitwo,D}}(A))$ 
where $\alpha$ ranges from 0 to 1 (depending on proportions between the 
numbers of distinct information parts \ione\  and \itwo).

\begin{table}
\caption{Multivariate MTE. The table MADEOF and its projection
{\it select distinct I,D from MADEOF} 
}
\label{tmultiv} 

\begin{center}
\begin{tabular}{| r | l || l | l |}
\multicolumn{4}{c}{ 
Decision table MADEOF}\\
\hline
 & \multicolumn{1}{c |}{I} 
 & \multicolumn{1}{c |}{D} 
&  \multicolumn{1}{c |}{D2}   \\
\hline
1.&ABD A.G.&center&wooden\\
2.&LQR Inc. & school&stone\\
  &LQR Inc. & school&wooden\\
3.&PTS Ltd.&center&stone\\
  &PTS Ltd.&center&wooden\\
  &PTS Ltd.&restaurant&stone\\
4.&XYZ Inc.&center&stone\\
5.&ZZZ Ltd. & restaurant&stone\\
  &ZZZ Ltd. & restaurant&wooden\\
  &ZZZ Ltd. & school    &wooden\\
\hline
\multicolumn{4}{l}{ 
m(\{(school,wooden)\})=1/5}\\
\multicolumn{4}{l}{ 
m(\{(school,stone)\})=2/5}\\
\end{tabular}
\begin{tabular}{| r | l || l |}
\multicolumn{3}{c}{ 
Projected onto I,D:}\\
\hline
 & \multicolumn{1}{c |}{I} &  \multicolumn{1}{c |}{D}   \\
\hline
1.&ABD A.G.&center\\
2.&LQR Inc. & school\\
3.&PTS Ltd.&center\\
  &PTS Ltd.&restaurant\\
4.&XYZ Inc.&center\\
5.&ZZZ Ltd. & restaurant\\
  &ZZZ Ltd. & school\\
\hline
\multicolumn{3}{r}{ 
m$^{\downarrow 
D}$(\{(school)\})=}\\
\multicolumn{3}{r}{ 
m(\{(school,stone)\})+}\\
\multicolumn{3}{r}{ 
m(\{(school,wooden)\})=3/ 5}
\end{tabular}
\end{center} 
\end{table}

\begin{table}
\caption{Variable Independence}
\label{tvi} 

\begin{center}
D3 - heating \\
\quad\\
\begin{tabular}{| r | l | l || l | l |}
\hline
 & \multicolumn{1}{c |}{I2} 
 & \multicolumn{1}{c |}{I3} 
 & \multicolumn{1}{c |}{D } 
& \multicolumn{1}{c |}{D3}  \\
\hline
1.&AAA GmbH &EC&school&electric\\
2.&AAA GmbH &GC&school&gas\\
3.&BBB Ltd. &EC&center&electric\\
  &BBB Ltd. &EC&restaurant&electric\\
4.&BBB Ltd. &GC&center&gas\\
  &BBB Ltd. &GC&restaurant&gas\\
5.&CCC Inc. &EC&center&electric \\
  &CCC Inc. &EC&restaurant&electric\\
6.&CCC Inc. &GC&center &gas\\
  &CCC Inc. &GC&restaurant&gas\\
\hline
\end{tabular}
\end{center} 

$$Bel = Bel^{\downarrow D}\oplus Bel^{\downarrow D2}$$
because the above table represents a cross product of the tables (without 
common columns) 

\begin{center}
\begin{tabular}{| r | l || l |}
\hline
 & \multicolumn{1}{c |}{I2} & \multicolumn{1}{c |}{D}  \\
\hline
1.&AAA GmbH&school\\
2.&BBB Ltd. &center\\
  &BBB Ltd.&restaurant\\
3.&CCC Inc.&center \\
  &CCC Inc.&restaurant\\
\hline
\end{tabular}
and
\begin{tabular}{| r | l || l |}
\hline
 & \multicolumn{1}{c |}{I3} & \multicolumn{1}{c |}{D3}  \\
\hline
1.&EC&electric\\
2.&GC&gas\\
\hline
\end{tabular}

\end{center} 
\end{table}
\begin{table}
\caption{Conditional Variable Independence}
\label{tcvi}

\begin{center}
I4 - painting company,
D4 - color,
D5 - finish,

\begin{tabular}{| r | l | l || l | l | l |}
\hline
 & \multicolumn{1}{c |}{I2} 
 & \multicolumn{1}{c |}{I4} 
 & \multicolumn{1}{c |}{D}  
 & \multicolumn{1}{c |}{D5}  
 & \multicolumn{1}{c |}{D4}  \\
\hline
1.&AAA GmbH &Messer&school     &wood     &green\\
  &AAA GmbH &Messer&school     &wood     &red\\
  &AAA GmbH &Messer&school     &plastic  &green\\
  &AAA GmbH &Messer&school     &plastic  &red\\
2.&BBB Ltd. &Messer&center     &metallic  &white  \\
  &BBB Ltd. &Messer&center     &metallic  &yellow \\
  &BBB Ltd. &Messer&center     &marble   &white  \\
  &BBB Ltd. &Messer&center     &marble   &yellow \\
3.&BBB Ltd. &Gabel &restaurant &wood     &red\\
4.&CCC Inc. &Messer&center     &metallic  &white  \\
  &CCC Inc. &Messer&center     &metallic  &yellow \\
  &CCC Inc. &Messer&center     &laminated&white  \\
  &CCC Inc. &Messer&center     &laminated&yellow \\
5.&CCC Inc. &Gabel &restaurant &plastic  &red\\
\hline
\end{tabular}
\end{center} 

In Bel of the above table variables D4 and D5 are conditionally independent 
given D in the sense of Shenoy's valuation-based systems because the above 
table is a relational join of the tables below (with D as a common column) 
{\tiny   
\begin{center}
\begin{tabular}{| r | l || l | l |}
\hline
 & \multicolumn{1}{c |}{I2} 
 & \multicolumn{1}{c |}{D}  
 & \multicolumn{1}{c |}{D5}  \\
\hline
1.&AAA GmbH &school     &wood\\
  &AAA GmbH &school     &plastic\\
2.&BBB Ltd. &center     &metallic\\
  &BBB Ltd. &center     &marble\\
  &BBB Ltd. &restaurant &wood\\
3.&CCC Inc. &center     &metallic\\
  &CCC Inc. &center     &laminated\\
  &CCC Inc. &restaurant  &plastic\\
\hline
\end{tabular}
\& 
\begin{tabular}{| r | l || l | l |}
\hline
 & \multicolumn{1}{c |}{I4} 
 & \multicolumn{1}{c |}{D } 
& \multicolumn{1}{c |}{D4}  \\
\hline
1.&Gabel &restaurant &red\\
2.&Messer&school     &green\\
  &Messer&school     &red\\
  &Messer&center     &white  \\
  &Messer&center     &yellow \\
\hline
\end{tabular}
\end{center} 
}
\end{table}

%

\subsection{Multivariate Beliefs and Multidecision Tables} 

We can extend our consideration to tables with multiple decision variables. 
In a straight forward way we can extend our definition of \DST 
measures to such 
tables and consider multivariate belief distributions (in all the decision 
variables). 
It is trivial to see that dropping a decision variables  does  not 
diminish the 
diversity of the information part. Hence dropping a decision variable D$i$ 
 from 
the set of decision variables {\bf D}  has the  same effect as dropping a 
variable in the belief function. That is for any set B of decision vectors in 
variables {\bf D}-\{D$i$\}: 
$m_{TAB ^{\downarrow {\tiny\bf D}-\{Di\} }} (B)=
m_{TAB} ^{\downarrow {\tiny\bf D}-\{Di\} }(B)= c \cdot
\sum_{A; B=A  ^{\downarrow {\tiny\bf D}-\{Di\} } } m(A)_{TAB}$
(c - normalizing factor)
See table \ref{tmultiv} for an example.

The operator of projection $\downarrow$ should be understood as the \DST 
projection operator applied to a belief function.

Let DTM be a decision table with decision variables D1 and D2. Let the 
information part consist of two disjoint parts I1 and I2. 
Let us consider the following views: 
\SQL{create view DTM1 (I1,D1) as select distinct I1,D1 from DTM;\\
     create view DTM2 (I2,D2) as select distinct I2,D2 from DTM;\\
     create view DTM12  as select distinct I1,I2,D1,D2 from DTM1,DTM2; } 
If now the table DTM12 is relationally identical with DTM, then we shall say 
that the decision variables D1 and D2 are independent in the decision table 
DTM.  
It is       not surprising that: 
$Bel_{DTM}=Bel_{DTM}^{\downarrow D1}\oplus Bel_{DTM}^{\downarrow D2}$.
This means that independence of decision variables in a decision table 
 implies 
independence of variables in the corresponding belief function. 
See table \ref{tvi}  for an example.

Let DTX be a decision table with decision variables D1, D2 and D3. Let the 
information part consist of two disjoint parts I1 and I2. 
Let us consider the following views: 
\SQL{create view DTX1 (I1,D1,D3) as select distinct I1,D1,D3 from DTX;\\
     create view DTX2 (I2,D2,D3) as select distinct I2,D2,D3 from DTX;\\
     create view DTX12  as select distinct I1,I2,D1,D2,D3 from DTX1,DTX2 
 where 
DTX1.D3=DTX2.D3; } 
If now the table DTX12 is relationally identical with DTX, then
we shall say 
that the decision variables D1 and D2 are independent given D3 in the 
 decision 
table DTX.  
It is       not surprising that: 
$Bel_{DTX}=Bel_{DTX}^{\downarrow D1,D3}\oplus Bel_{DTX}^{\downarrow D2,D3}$.
But this means that the variables D1 and D2 are independent given D3 in the 
belief function $Bel_{DTX}$ in the sense of Shenoy's VBS \cite{Shenoy:94}. 
See table \ref{tcvi} for an example.

These results mean that analysis of independence and conditional independence 
of variables in a belief function corresponding to a decision table may serve 
as an indicator of presence or absence of independence or multivalued 
dependence in the decision table.

\section{Concluding Remarks}

A novel case-based interpretation of \DST belief 
functions which is qualitative in nature and has the potential to represent a 
number of \DST operations has been presented. The interpretation is based on 
 rough sets (in 
connection with decision tables), but differs from previous interpretations 
 of 
this type e.g. 
\cite{Skowron:94} in that it counts the diversity rather than frequencies in 
the decision table. The interpretation has the property that 
given a 
definition of the \DST measure of objects in the interpretation domain 
(decision table) 
we can perform operations in the interpretation domain  (e.g. combining 
decision tables) and the  measure  of  the  resulting   object  is 
derivable from 
measures of component objects via \DST operator (e.g. combination). We 
demonstrated this property for Dempster rule of combination, marginalization, 
Shafer's conditioning, 
independent variables, 
Shenoy's notion of conditional independence of variables. 
Other known case-based (frequency or probabilistic) interpretations fall 
 short 
of this property. E.g. in \cite{Skowron:94} complex rational number 
arithmetic, unnatural for decision tables, is needed to achieve 
 compatibility 
of the final decision table with Dempster rule of combination.
In \cite{Kyburg:87} (probabilistic interpretation) only lower and upper 
 bounds 
are found for Dempster rule. 
In \cite{Fagin:91} (probability structures interpretation)
the belief function obtained from 
 the  Dempster rule 
 is a potential, but not necessary result
of the corresponding 
probabilistic structure combination operation. 
 See also \cite{Smets:92} 
for discussion of other interpretations. 

As probabilistic (frequency based) models          of 
MTE seem to fail in general, we looked for alternatives. 
The result of any  experiment with multiple outcomes
 can evaluated along two dimensions: 
the quantitative and the qualitative one.
In most real life cases we are more interested in the quantitative aspects.
Sometimes, however, the qualitative side may be of more interest.
E.g. legal applications we would treat  suspiciously similar
testimonies as a single argument in favour or against a hypothesis: we would rely on the count of diversity of arguments rather than on their actual counts. 
And MTE was claimed to be applicable just in legal reasoning
\cite{Shafer:90ijar}. 
Therefore we considered just the diversity  as a possible alternative for a 
model of MTE and this idea turned out to very fruitful. 
 The diversity is well handled by relational model of databases. Though at a first glance the count of cases
in a relational database may appear identical with counting frequencies of objects / events, the difference starts as soon as we make a projection on the subset of attributes. It turns out that projected frequencies 
differ significantly from the counts of cases in a projected relational database.  Therefore we say that our approach is non-frequency, non-quantitative, that is qualitative one.

The paper presents SQL statements performing the calculations of the \DST 
measures and the \DST related operations on the decision tables. 

The new interpretation  may be directly applied in the domain of multiple 
decision decision tables: independence of decision variables or Shenoy's 
conditional independence in the sense of \DST may serve as an indication of 
possibility of decomposition of the decision table into smaller but 
 equivalent 
tables. 
Furthermore it may 
be applied in the area of Cooperative Query Answering \cite{Ras:91}.  
The problem there is that  a query posed  to  a local relational database 
system may contain an unknown attribute. But possibly other co-operating 
db systems know it and may explain it to the queried system in terms of known 
attributes, shared by the various systems. The uncertainties studied in the 
decision tables arise here in a natural way and our interpretation  may be 
used to measure these uncertainties in terms of \DST (as diversity of 
support). Furthermore, if several co-operating systems respond, then the 
queried system may calculate the overall uncertainty measure using \DST 
combination of measures of individual responses.

Now we can ask how to understand then a \DST belief function in the light of 
our experience. One possibility is to consider the belief function as a 
measure of diversity of support.  This is an obvious departure from frequency 
interpretations proposed by Shafer and others. No mater how frequently the 
same piece of evidence is presented, it is counted once. 
This insight may encourage to revise other known interpretations of \DSTNB.  
In the "legal" 
interpretations e.g. \cite{Shafer:90ijar}, the witnesses in  favor of a 
hypotheses should be counted 
separately, if their statements differ in unimportant details permitting to 
deduce that their statements are personal and not studied in. In the 
"probability of provability" approach \cite{Pearl:88}
 not a probability of correctness, but 
rather the number of distinct valid proofs of a statement  should be counted. 
In the "possible world semantics" \cite{Ruspini:86}
 the worlds should not be assigned a 
probability, but rather distinct possible worlds should be counted that 
differ in non-essential details. 
Then the operation of combination of independent evidence in the "legal" 
interpretation is just mixing compatible statements of two sets of 
witnesses 
(which saw the same event from different perspective) and counting different 
possible combinations. 
    {
In the "probability of provability" approach the 
combination would  mean  putting  together  conclusion  compatible 
proofs stemming 
from distinct domains (e.g. macro and micro-physical observations) and 
counting legal combinations. In the possible worlds semantics we may combine 
worlds spanned over disjoined sets of dimensions. 
}

Please notice also that the rough-set interpretation sheds some light onto 
what the 
concept of "evidence" may mean.  The "evidence" are just 
different sets of information attributes and the "independence" means 
(deterministic) unpredictability of attribute values of the one set from the 
other set.  This should not be confused with predictability of the decision 
variable. Nor with stochastic predictability which may be present. 
In the "legal" interpretation, independence would be measured with 
non-predictability of insignificant details. 
    {
In the "provability" 
interpretation the independence may be measured by mutual non-derivability of 
the sets of underlying axioms.   In the possible world semantics by 
possibility of putting together projections onto separated sets of dimension 
axes.}  

Further studies  on interpreting other known \DST operators in the spirit of 
qualitative  interpretation presented in this paper are needed and may reveal 
new potential applications of \DST to real world problems.

\newcommand{\eng}[1]{}    
\newcommand{\pol}[1]{}    
\newcommand{\deu}[1]{}    
\newcommand{\A}[2]{#1 #2} 
\newcommand{\Ed}[2]{#2 #1} 
\newcommand{\BT}[1]{{\it #1}. } 
\newcommand{\JT}[1]{{\it #1} } 
\newcommand{\AT}[1]{#1. } 
\newcommand{\VY}[2]{{\bf #1}(#2)} 

\renewcommand{\eng}[1]{#1}    
\newcommand{\IN}{\pol{[w:]}\deu{[in:]}\eng{In:}} 
\newcommand{\LitStelle}[2]{\bibitem{#1} }   

\newcommand{\ReadingsIn}{
\Ed{G.}{Shafer}, \Ed{J.}{Pearl},eds, \BT{Readings in Uncertain 
Reasoning} 
Morgan Kaufmann Pub.
 Inc., San Mateo CA, 
 (1990)}

{\it \noindent
 Institute of Computer Science\\ Polish Academy of Sciences\\ 
01-237 Warszawa,  ul. Ordona 21, 
Poland \\
e-mail: klopotek,stw@ipipan.waw.pl}

\end{document}